\documentclass[final]{cvpr}

\usepackage{times}
\usepackage{epsfig}
\usepackage{graphicx}
\usepackage{amsmath}
\usepackage{amssymb}

\usepackage{url}            
\usepackage{booktabs}       
\usepackage{amsfonts}       
\usepackage{nicefrac}       
\usepackage{microtype}      
\DeclareMathOperator*{\argmax}{argmax}
\usepackage{multirow}
\usepackage{multicol}
\usepackage{makecell}
\usepackage{soul}
\usepackage{color}
\usepackage[dvipsnames]{xcolor}

\usepackage[pagebackref=true,breaklinks=true,colorlinks,bookmarks=false]{hyperref}

\def\confYear{CVPR 2021}

\begin{document}

\title{WOAD: Weakly Supervised Online Action Detection in Untrimmed Videos}

\author{Mingfei Gao, Yingbo Zhou, Ran Xu, Richard Socher, Caiming Xiong\\
Salesforce Research\\
{\tt\small \{mingfei.gao,yingbo.zhou,ran.xu,rsocher,cxiong\}@salesforce.com}
}

\maketitle

\begin{abstract}
Online action detection in untrimmed videos aims to identify an action as it happens, which makes it very important for real-time applications. Previous methods rely on tedious annotations of temporal action boundaries for training, which hinders the scalability of online action detection systems. We propose WOAD, a weakly supervised framework that can be trained using only video-class labels. WOAD contains two jointly-trained modules, i.e., temporal proposal generator (TPG) and online action recognizer (OAR). Supervised by video-class labels, TPG works offline and targets at accurately mining pseudo frame-level labels for OAR. With the supervisory signals from TPG, OAR learns to conduct action detection in an online fashion. Experimental results on THUMOS'14, ActivityNet1.2 and ActivityNet1.3 show that our weakly-supervised method largely outperforms weakly-supervised baselines and achieves comparable performance to the previous strongly-supervised methods. Beyond that, WOAD is flexible to leverage strong supervision when it is available. When strongly supervised, our method obtains the state-of-the-art results in the tasks of both online per-frame action recognition and online detection of action start.
\end{abstract}

\section{Introduction}
Temporal Action Localization aims to detect temporal action boundaries in long, untrimmed videos. Most previous methods are under offline settings~\cite{buch2017sst,chao2018rethinking,dai2017temporal,gao2017turn,shou2016temporal,zhao2017temporal}, where they can observe the entire action before making decisions. However, applications such as surveillance systems and autonomous cars, are required to interact with the world in real time based on their accumulative observations up to now. Online Action Detection~\cite{de2016online} is proposed to address this problem, where methods need to identify occurring actions moment-to-moment without access to future information. With different focuses, recent online action detectors consider two sub-tasks: (1) online per-frame action recognition~\cite{de2016online,gao2017red,xu2019trn} and (2) online detection of action start~\cite{gao2019startnet,shou2018online}. The former task focuses on the general capability of recognizing the action category of each coming frame. On the other hand, detecting action starts in a timely manner is more important to some real-world applications. For example, an autonomous car needs to recognize ``line merging'' of another vehicle as soon as it starts. While, it is challenging to detect action starts due to the similar appearances near the start points and the lack of training data. The later task specially targets on this problem. Our method jointly addresses these two tasks. 
\begin{figure}[t]
    \centering
    \includegraphics[width=1.0\linewidth]{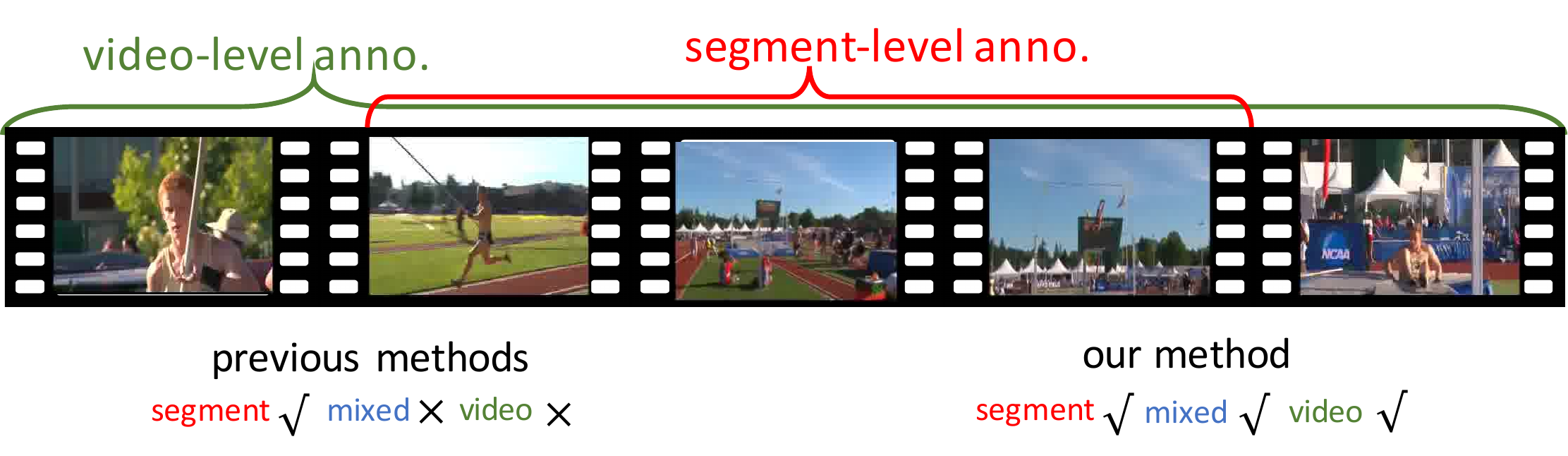}
    \caption{Comparison between previous methods and our approach. Previous methods require \textcolor{red}{segment-level} annotations (start and end times of actions) during training, which leads to high human labeling cost. In contrast, our method can be trained with \textcolor{OliveGreen}{video-level} annotations (video-level action labels) and is flexible to utilize (\textcolor{red}{full}/ \textcolor{blue}{partial}) segment-level supervision when it is available.}
    \label{fig: idea}
\end{figure}

Although previous methods have achieved promising progress, they rely on segment-level annotations of action boundaries for training (see Fig.~\ref{fig: idea}). However, annotating action boundaries in long, untrimmed videos involves possibly ambiguous decisions and requires significant amount of human labor. This hinders the scalability of model learning, particularly for videos embodying complex semantics. Compared to the segment-level boundaries, video-level action classes are much easier to acquire. With the help of text-based video retrieval techniques, video-class labels may be obtained almost for free from the internet at a large scale.

\begin{figure*}[htbp]
    \centering
    \includegraphics[width=1.0\linewidth]{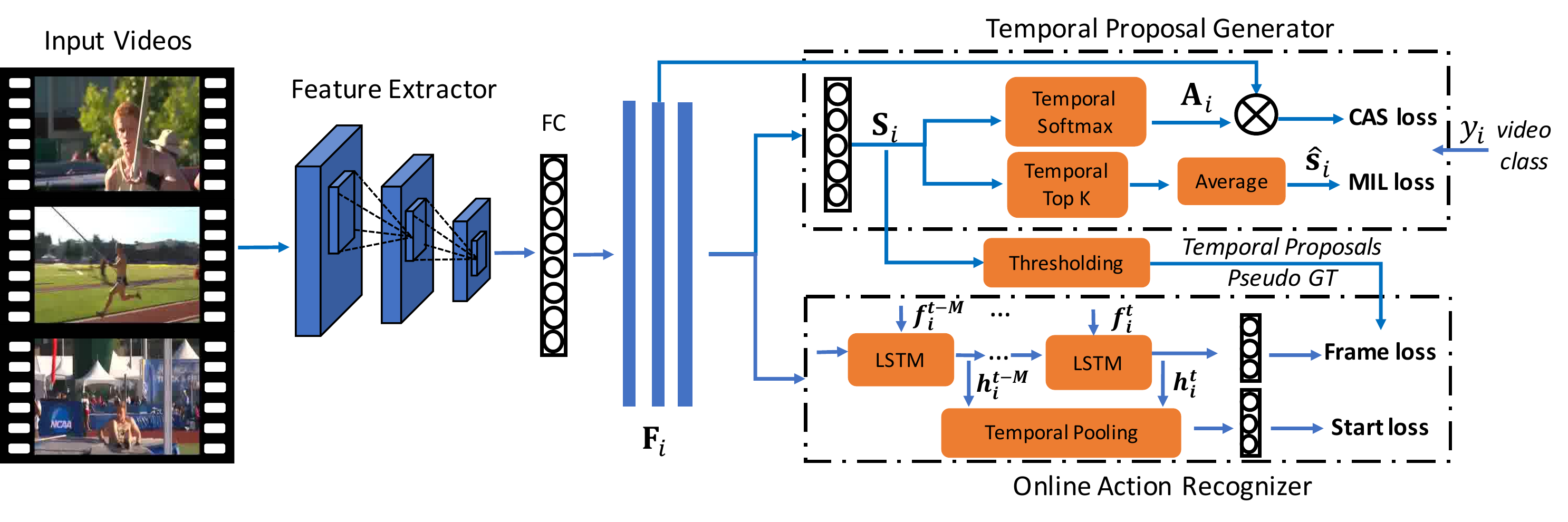}
    \caption{Illustration of the proposed WOAD in the training phase. A feature extractor is used to extract features of the input video. Frame features, $\textbf{F}_i$, are then obtained by a fully connected layer (FC) activated with ReLU and serve as inputs to both Temporal Proposal Generator (TPG) and Online Action Recognizer (OAR). TPG is trained using video-class labels and its generated class-wise temporal proposals are used as pseudo ground truth of action boundaries to supervise the training of OAR. See details in Sec.~\ref{sec: tpg} and Sec.~\ref{sec: oar}.
    }
    \label{fig: pipeline}
\end{figure*}

To take advantage of the easy-to-obtain video-level annotations, we propose WOAD, a \textbf{W}eakly supervised \textbf{O}nline \textbf{A}ction \textbf{D}etection framework, that \emph{can be trained with video-class labels only}. Detecting actions using weak supervision in an online scenario is challenging since (1) online action detectors generally require per-frame labels for training, so it is hard to utilize video-level labels as supervision and (2) it is not trivial for a model to be accurate for action recognition and sensitive to action starts without access to future information. As shown in Fig.~\ref{fig: pipeline}, our proposed WOAD contains two jointly-trained modules, i.e., Temporal Proposal Generator (TPG) and Online Action Recognizer (OAR), each of which focuses on handling one of the challenges. Supervised by video-class labels, TPG mines class-wise temporal action proposals that can be used as pseudo per-frame labels for OAR. While, OAR aims at conducting both per-frame action recognition and start detection jointly in an online fashion. 

The proposed design has the following benefits: (1) TPG is used only during training for pseudo labels generation, so it can fully utilize temporal relation of frames (e.g. grouping nearby frames of the same class to improve proposal generation) without online constraint; (2) the design of OAR directly targets at improving the online tasks without being distracted by the weakly supervised setting and (3) the joint training could help learn better representations.

Our contributions are summarized as follows: (1) we introduce a novel method for weakly supervised online action detection. To the best of our knowledge, this is the first work that addresses the problem using weak supervision; (2) our method is flexible to combine weak and strong supervision when only a part of videos have strong annotations and (3) experimental results show that our method largely outperforms weakly-supervised baselines and achieves comparable performance to the previous strongly-supervised methods. When strongly supervised, our method obtains the state-of-the-art results in the tasks of both online per-frame action recognition and online detection of action start.
\section{Related Work}

\textbf{Temporal Action Localization}.
The goal of temporal action detectors is to find the temporal boundaries of actions in untrimmed videos. Most existing methods work in offline settings, where they can make predictions after processing the entire actions. Shou et al. proposed S-CNN~\cite{shou2016temporal} to conduct action proposal generation, classification and regression via multi-stage networks. Dai et al. introduced TCN~\cite{dai2017temporal} that utilizes temporal context of proposals to improve proposal generation. Xu et al. presented R-C3D~\cite{xu2017r} that improves model efficiency by sharing the processing stages of proposal generation and classification. Buch et al. proposed SST~\cite{buch2017sst} to conduct fast proposal generation. Zeng et al. modeled relations among proposals using Graph Convolutional Networks~\cite{kipf2016semi} and improved feature representations in~\cite{zeng2019graph}.

\textbf{Online Action Detection}.
Online action detectors identify the occurring action in untrimmed, streaming videos based on the past and current observations. Geest et al. first posed this problem as online per-frame action recognition and set up several baselines and evaluation metrics in~\cite{de2016online}. Following this direction, Gao et al. introduced RED~\cite{gao2017red} which conducts current and future action predictions jointly. Xu et al. proposed TRN~\cite{xu2019trn} that uses the predicted future actions to improve action recognition at the current time. Eun et al. introduced IDU to accumulate input information based on its relevance to the current action in~\cite{eun2020idu}. Compared to per-frame action recognition, online detecting action starts is more important for some applications and is more challenging due to the similar appearance near starts and the lack of training data. Shou et al. first proposed an online framework in~\cite{shou2018online} and treated the problem as a classification task. Gao et al. presented StartNet~\cite{gao2019startnet} which set the new state-of-the-art performance. However, these methods depend on the annotations of action boundaries for training and are evaluated on either per-frame action recognition or action start detection. Our work jointly handles these two tasks using weak supervision.

\textbf{Weakly Supervised Offline Action Detection}.
Extensive studies have been done in offline action detection with video-class lables as supervision. Wang et al. introduced UntrimmedNet~\cite{wang2017untrimmednets} to model actions from untrimmed videos. Shou et al. improved UntrimmedNet by introducing Outer-Inner-Contrastive loss~\cite{shou2018autoloc}. Using only video-level labels, W-TALC~\cite{paul2018w} learns action representations using MIL and co-activity similarity losses. Liu et al. focused on the completeness of actions in~\cite{Liu_2019_CVPR} and BasNet~\cite{lee2020background} improved weakly supervised action localization by background suppression. In~\cite{narayan20193cnet}, Narayan et al. optimized the models by jointly minimizing category, count and center losses. Yuan et al. proposed MAAN~\cite{yuan2018marginalized} to relieve the effect of the dominant response of the most salient regions. Luo et al. explicitly modeled the key instances assignment via a EM-MIL approach in~\cite{luo2020weakly}. Nguyen et al. learned a rich notion of actions via background modeling in~\cite{nguyen2019weakly}. Min et al. proposed A2CL-PT~\cite{min2020adversarial} to learn discriminative features and distinguish background. Existing offline methods are not well suitable in the online setting since (1) offline methods target at predicting action segments that have significant temporal overlap with ground-truth, so they are not designed to be good at per-frame recognition and are not sensitive to start; (2) most of them adopt temporal prediction grouping
strategy to improve performance during inference which violates online constraint and (3) technically, they prefer feedforward networks while online
action detectors work better with RNNs.
\section{Weakly Supervised Online Action Detection}
\subsection{Framework Overview}
For online action detection, the input to the system is a streaming, untrimmed video, $\textbf{V}_i$, represented as a sequence of image frames $[I_i^1, I_i^2,...,I_i^{T_i}]$, where $i$ denotes video index and $T_i$ is video length. At each time $t$, the system takes $I_i^t$ as input. It predicts, $\textbf{a}_i^t $, the probability of the current action category (online per-frame action recognition) and determines, $\textbf{as}_i^t$, the probability that an action start occurs (online detection of action start). Under the constraint of online setting, no future information is available in the inference phase. Previous approaches require annotations of temporal action boundaries for training. The proposed method can be trained using only video-class labels.

As shown in Fig.~\ref{fig: pipeline}, our method contains two modules, i.e., the Temporal Proposal Generator (TPG) and the Online Action Recognizer (OAR). During training, TPG is supervised by video-class labels and outputs class-wise temporal proposals (Sec.~\ref{sec: tpg}). The proposals serve as pseudo ground truth of action boundaries which can be used as per-frame labels to supervise the training of OAR (Sec.~\ref{sec: oar}). During inference, only OAR is used for online action detection.

\subsection{Temporal Proposal Generator}
\label{sec: tpg}
There are different options to implement our Temporal Proposal Generator (TPG). In this work, we focus on utilizing Multiple Instance Learning (MIL) loss and Co-Activity Similarity (CAS) loss proposed in~\cite{paul2018w}. Next, we will revisit the definitions of these two losses

Let $\textbf{F}_i = [\textbf{f}_i^1, \textbf{f}_i^2,..., \textbf{f}_i^{T_i}]^\top \in \mathbb{R}^{T_i\times D}$ indicates the features of $\textbf{V}_i$ just before the TPG module as shown in Fig.~\ref{fig: pipeline}, where $\textbf{f}_i^t$ indicates the feature of the frame at time step $t$, and $T_i$ denotes the number of frames in $\textbf{V}_i$. We obtain per-frame scores, $\textbf{S}_i = [\textbf{s}_i^1,  \textbf{s}_i^2,..., \textbf{s}_i^{T_i}]^\top \in \mathbb{R}^{T_i\times C}$ by projecting the features to action class space and $ \textbf{s}_i^t = [s_{i1}^t, s_{i2}^t,...,s_{iC}^t] \in \mathbb{R}^{C}$ indicates scores of frame $t$ over $c=[1,2,...,C]$ classes. For each class $c$, a video-level score, $\hat{s}_{ic}$, is obtained by averaging over the top $K_i$ frame scores as $\hat{s}_{ic}=\frac{1}{K_i} \sum_{t\in \mathbb{K}_{ic}} s_{ic}^t$, where $\mathbb{K}_{ic}$ indicates the set of top $K_i$ frames for class $c$ over $T_i$ frames, $K_i=\lfloor\frac{T_i}{\kappa}\rfloor$ and $\kappa$ is fixed to be 8.

\textbf{MIL loss}, $L_{MIL}$, is defined as the cross entropy loss between the video-class label, $y_{i}$, and the predicted video-class probability, $p_{i}$, where $p_{i}$ is obtained by applying softmax in classes over $\hat{\textbf{s}}_i=[\hat{s}_{i1}, \hat{s}_{i2},...,\hat{s}_{iC}]$.

\textbf{CAS loss} encourages regions of videos containing similar activities to have similar feature representations, and those containing different activities to have different representations. High- and low-attention region feature representations, $\Psi$ and $\Phi$, are introduced to achieve this goal. For class $c$, $\Psi_{ic} = \textbf{F}_i^\top\textbf{A}_{ic}$ and $\Phi_{ic} = \frac{1}{T_i-1}\textbf{F}_i^\top(1-\textbf{A}_{ic})$, where $\textbf{A}_{ic} \in \mathbb{R}^{T_i}$ is a temporal attention vector, obtained by applying temporal softmax over frame scores, $\textbf{S}_i$ 

Intuitively, $\Psi_{ic}$ aggregates features of regions with high probability containing the activity, while  $\Phi_{ic}$ aggregates those of regions that are unlikely involving in the activity. For class $c$, a positive video pair, $\textbf{V}_i$ and $\textbf{V}_j$, is constructed if $y_{ic}=y_{jc}=1$. Their pair-wise loss is calculated as 
\begin{equation}
    \begin{split}
   &\frac{1}{2}\{\max(0, d(\Psi_{ic}, \Psi_{jc})-d(\Psi_{ic}, \Phi_{jc})+\delta)\\
   &+\max(0, d(\Psi_{ic}, \Psi_{jc})-d(\Phi_{ic}, \Psi_{jc})+\delta)\}, 
   \end{split}
\end{equation}
where $d(x,z)$ denotes cosine similarity of $x$ and $z$, and $\delta$ is a margin parameter. CAS loss, $L_{CAS}$, is the average loss over all positive video pairs of all classes in the training batch.

\textbf{Proposal generation} is conducted via a two-stage thresholding strategy. First, a threshold, $\theta_{class}$, is used to discard categories having small video-level confidence scores. Then, a second threshold, $\theta_{score}$, is applied on the frame scores of the remaining categories, $s_{ic}^t$, along the temporal axis. Taking advantage of temporal constraint of frames, nearby frames with the same category are grouped to obtain the class-wise temporal proposals and action starts are thus obtained. After that, the video-class labels are used to filter out the proposals with wrong categories.

\subsection{Online Action Recognizer}
\label{sec: oar}
Online Action Recognizer (OAR) sequentially takes $\textbf{f}_i^t$ as input and outputs per-frame action scores over classes including background, $\textbf{a}_i^t \in \mathbb{R}^{(C+1)}$, and a class-agnostic start score, $\textbf{st}_i^t \in \mathbb{R}^{2}$, indicating the probabilities of this frame being a start point or not. 

Our OAR is constructed by a LSTM with temporal pooling. The LSTM updates its hidden and cell states, $\textbf{h}_i^t$ and $\textbf{c}_i^t$, at each time step as 
\begin{equation}
    \textbf{h}_i^t, \textbf{c}_i^t = LSTM(\textbf{h}_i^{t-1},  \textbf{c}_i^{t-1}, \textbf{f}_i^t).
\end{equation}

Then, $\widetilde{\textbf{h}}_i^t$ is obtained by applying max pooling along temporal axis from $\textbf{h}_i^{t-M}$ up to $\textbf{h}_i^t$ as in Eq.~\ref{eq: temppool}, where $M$ indicates the length of the temporal window. 

\begin{equation}
    \label{eq: temppool}
    \widetilde{\textbf{h}}_i^t = max \ pool(\textbf{h}_i^{t-M}, \textbf{h}_i^{t-M+1},...,\textbf{h}_i^t).
\end{equation}

$\textbf{a}_i^t$ and $\textbf{st}_i^t$ are obtained by a linear projection followed by the softmax operation on $\textbf{h}_i^t$ and $\widetilde{\textbf{h}}_i^t$, respectively as in Eq.~\ref{eq: label_project}, where $\textbf{W}_a$ and $\textbf{W}_{st}$ indicate the parameters of the classifiers.

\begin{equation}
\label{eq: label_project}
    \begin{aligned} 
    \textbf{a}_i^t &= softmax(\textbf{W}_a^{\top} \ \textbf{h}_i^t).  \\ 
    \textbf{st}_i^t &= softmax(\textbf{W}_{st}^{\top} \ \widetilde{\textbf{h}}_i^t).
    \end{aligned}
\end{equation}

In each training batch, we convert the proposal boundaries of each class $c$ (obtained from TPG) to per-frame action labels, $l_{jc}$ and binary start labels, $\zeta_{jm}$, where $j=\{1,2,..,\widetilde{T}\}$ indicates the index of a frame, $\widetilde{T}$ is the total number of frames in the training video batch and $m\in\{0,1\}$ differentiates the non-start and start. We use cross entropy loss between $l_{jc}$ and the predicted action probability, $a_{jc}$, to form frame loss and utilize focal loss~\cite{lin2017focal} between $\zeta_{jm}$ and the predicted start probability, $st_{jm}$, to construct start loss as shown in Eq.~\ref{eq: oar_loss}, where $\gamma$ is a hyper parameter.

\begin{equation}
\label{eq: oar_loss}
\begin{split}
    L_{OAR} \ = \ & \underset{frame \ loss}{\underbrace{- \frac{1}{\widetilde{T}}\sum_{j=1}^{\widetilde{T}}\sum_{c=0}^{C}\ l_{jc}\log a_{jc}}}  + \\ 
    &\underset{start \ loss}{\underbrace{\ -  \frac{1}{\widetilde{T}} \sum_{j=1}^{\widetilde{T}} \sum_{m=0}^{1}\ \zeta_{jm} \ (1-st_{jm})^\gamma \log st_{jm}}}.
\end{split}
\end{equation}
\subsection{Model Optimization and Inference}

\textbf{Optimization}. Our Temporal Proposal Generator (TPG) and Online Action Recognizer (OAR) are jointly optimized by minimizing
\begin{equation}
    \label{eq: loss}
    L_{total}=L_{OAR}+\lambda L_{TPG},
\end{equation}

where $L_{TPG}=L_{MIL}+L_{CAS}$. $L_{MIL}$ is computed for each videos and $L_{CAS}$ is calculated using the positive video pairs in the training batch. Each video is segmented to non-overlapping training sequences which are used to calculate $L_{OAR}$. As shown in Fig.~\ref{fig: pipeline}, proposals for OAR supervision are continuously updated. To reduce computation, we update the proposals every $N$ training iterations.

\textbf{Inference}. For the online action detection tasks, only OAR is used during inference. Proceeding sequentially, OAR outputs $\textbf{a}_i^t$ and $\textbf{st}_i^t$ at each time step $t$. $\textbf{a}_i^t$ can be used directly as the per-frame action prediction. Following~\cite{gao2019startnet}, scores of action starts, are obtained by $\textbf{as}_{i(1:C)}^t=\textbf{a}_{i(1:C)}^t*\textbf{st}_{i1}^t$ and $\textbf{as}_{i0}^t = \textbf{a}_{i0}^t*\textbf{st}_{i0}^t$, where $(1:C)$ indicates positive classes and $(0)$ denotes background. Then, we generate action starts following the criteria~\cite{shou2018online,gao2019startnet}: (1) the predicted class $\hat{c}_i^t = \underset{c}{\argmax} (\textbf{as}_i^t)$ is an action; (2) the maximum action score $\textbf{as}_{i\hat{c}_i^t}^t$ exceeds a threshold (set to be 0) and (3) $\hat{c}_i^t \neq \hat{c}_i^{t-1} $. As indicated, $\textbf{st}_i^t$ is used to boost the scores if a start is predicted at time $t$ and suppress those otherwise.

\begin{table*}[ht]
    \centering
    \begin{tabular}{l|c||c|c|c|c|c|c|c|c|c|c||c}
        \multicolumn{12}{c}{\ \ \ \ \ \ \ \ \ \ \ \ \ \ \ \ \ \ \ \ \ \ \ \ \ \ \ \ \ \ \ \ \ \ \ \ \ \ \ \ \ \ \ \ \ mean P-AP@ Time Threshold (Seconds)} \\
        Methods & Sup. & 1 & 2 & 3 & 4 & 5 & 6 & 7 & 8 & 9 & 10 & mean F-AP\\
        \midrule
        W-TALC~\cite{paul2018w}& \multirow{2}{*}{V}& 16.2& 26.0& 31.3& 34.6& 36.2& 37.6& 38.6& 39.3& 39.9&  40.3& 48.0\\
        \cmidrule{1-1} \cmidrule{3-13}
        \textbf{WOAD} & & \textbf{21.9} & \textbf{32.9} & \textbf{40.5} & \textbf{44.4} &\textbf{48.1} & \textbf{49.8} &\textbf{50.8}&\textbf{51.7} &\textbf{52.4} & \textbf{53.1} & \textbf{54.4}\\
    \end{tabular}
    \caption{Comparison with our baseline under weakly supervised setting on THUMOS'14. V indicates video-level (weak) supervision.}
    \label{tab: thumos_weak_compare}
\end{table*}

\begin{table*}[ht]
    \centering
    \begin{tabular}{l|c||c|c|c|c|c|c|c|c|c|c||c}
        \multicolumn{12}{c}{\ \ \ \ \ \ \ \ \ \ \ \ \ \ \ \ \ \ \ \ \ \ \ \ \ \ \ \ \ \ \ \ \ \ \ \ \ \ \ \ \ \ \ \ \ mean P-AP@ Time Threshold (Seconds)} \\
        Methods & Sup. & 1 & 2 & 3 & 4 & 5 & 6 & 7 & 8 & 9 & 10 & mean F-AP\\
        \midrule
        W-TALC~\cite{paul2018w}& \multirow{2}{*}{V}& 5.2& 8.5& 10.7& 12.8& 14.5& 15.9&17.1 &18.1 &19.1 &20.1  &53.8 \\
        \cmidrule{1-1} \cmidrule{3-13}
        \textbf{WOAD} & &\textbf{7.9}&\textbf{11.6}&\textbf{14.3}&\textbf{16.4}&\textbf{18.8}&\textbf{20.3}&\textbf{22.2}&\textbf{23.4}&\textbf{24.7}&\textbf{25.3}& \textbf{66.7}\\
    \end{tabular}
    \caption{Comparison with our baseline under weakly supervised setting on ActivityNet1.2. V indicates video-level (weak) supervision.}
    \label{tab: activity1.2_weak_compare}
\end{table*}

\begin{table*}[ht]
    \centering
    \begin{tabular}{l|c||c|c|c|c|c|c|c|c|c|c}
        \multicolumn{12}{c}{\ \ \ \ \ \ \ \ \ \ \ \ \ \ \ \ \ \ \ \ \ \ \ \ \ \ \ \ \ \ \ \ \ \ \ mean P-AP@ Time Threshold (Seconds)} \\
        
        Methods & Sup. & 1 & 2 & 3 & 4 & 5 & 6 & 7 & 8 & 9 & 10 \\
        \midrule
        StartNet~\cite{gao2019startnet}& S
        &\underline{21.9} &\underline{33.5}  &39.6  &42.5  &46.2  &46.6  &47.7  &48.3  & 48.6 &49.0  \\
        \midrule
        \multirow{2}{*}{\textbf{WOAD}} & V& \underline{21.9} & 32.9 & \underline{40.5} & \underline{44.4} &\underline{48.1} & \underline{49.8} &\underline{50.8}&\underline{51.7} &\underline{52.4} & \textbf{53.1} \\
        \cmidrule{2-12}
        \cmidrule{2-12}
        & S & \textbf{28.0}&\textbf{40.6} &\textbf{45.7}&\textbf{48.0}&\textbf{50.1}&\textbf{51.0}&\textbf{51.9}&\textbf{52.4}&\textbf{53.0}&\textbf{53.1} \\
    \end{tabular}
    \caption{Comparison with strongly-supervised method for online detection of action start on THUMOS'14. V and S denote video-level (weak) and segment-level (strong) supervision, respectively.  \textbf{Best} and \underline{second-best} per column are highlighted.}
    \label{tab: thumos_start}
\end{table*}
\begin{table*}[ht]
    \centering
    \begin{tabular}{l|c||c|c|c|c|c|c|c|c|c|c}
        \multicolumn{12}{c}{\ \ \ \ \ \ \ \ \ \ \ \ \ \ \ \ \ \ \ \ \ \ \ \ \ \ \ \ \ \ \ \ \ \   mean P-AP@ Time Threshold (Seconds)} \\
        Methods & Sup. & 1 & 2 & 3 & 4 & 5 & 6 & 7 & 8 & 9 & 10 \\
        \midrule
        StartNet~\cite{gao2019startnet}&S &7.5 & 11.5 & 14.1 & \underline{16.5} & 18.4 &  19.7 &20.9 &21.8 & 22.9 & 23.6\\
        \midrule
        \multirow{2}{*}{\textbf{WOAD}} & V &\underline{7.9}&\underline{11.6}&\underline{14.3}&16.4&\underline{18.8}&\underline{20.3}&\underline{22.2}&\underline{23.4}&\underline{24.7}&\underline{25.3}  \\
        \cmidrule{2-12}
        \cmidrule{2-12}
        & S &\textbf{8.7}& \textbf{13.6}& \textbf{17.0}&\textbf{19.7}&\textbf{21.6}&\textbf{23.0}&\textbf{24.7}&\textbf{25.8}&\textbf{26.8}&\textbf{27.7}  \\
    \end{tabular}
    \caption{Comparison with strongly-supervised method for online detection of action start on ActivityNet1.2. V and S denote video-level (weak) and segment-level (strong) supervision, respectively. \textbf{Best} and \underline{second-best} per column are highlighted.}
    \label{tab: anet_start}
\end{table*}
\begin{table}[ht]
    \centering
    \begin{tabular}{l|c||c|c}
        Methods & Sup. & mean F-AP & mean P-AP@ 1\\
        \midrule
        TRN~\cite{xu2019trn}& \multirow{2}{*}{S}& 43.8& -- \\
        StartNet~\cite{gao2019startnet}& & -- & 4.9 \\
        \midrule
        \multirow{2}{*}{\textbf{WOAD}}& V & 44.0& 5.1\\
        \cmidrule{2-4}
        & S & \textbf{46.8}& \textbf{5.5}  \\
    \end{tabular}
    \caption{Comparison with strongly-supervised methods on ActivityNet1.3. V and S denote video-level (weak) and segment-level (strong) supervision.}
    \label{tab: anet_1.3}
\end{table}

\begin{table}[ht]
    \centering
    \begin{tabular}{l|c||c|c|c}
        Methods & Sup. & Param. $\#$ & Infer time& mean F-AP\\
        \midrule
        TRN~\cite{xu2019trn}& S
        &314M &2.60 ms &51.0\\
        \midrule
        \textbf{WOAD} & V &110M&0.40 ms&\textbf{54.4}\\
    \end{tabular}
    \caption{Comparison with strongly-supervised method for online per-frame action recognition on THUMOS'14. The reported times do not include the processing time of feature extraction. V and S indicate video-level (weak) and segment-level (strong) supervision.}
    \label{tab: thumos_frame}
\end{table}
\begin{table*}[ht]
    \centering
    \begin{tabular}{l|c|c|c|c|c|c|}
    Methods $\rightarrow$ & TRN~\cite{xu2019trn}& \multicolumn{5}{c|}{\textbf{WOAD}}\\
    Supervision $\rightarrow$ &S & V&V+$10\%$S&V+$30\%$S& V+$50\%$S& S \\
    \midrule
    mean F-AP $\rightarrow$ &51.0&54.4&55.0& 59.3& 62.6&\textbf{67.1} \\
    \end{tabular}
    \caption{Comparison with strongly-supervised SOTA method on THUMOS'14. V+$10\%$S means that $10\%$ of videos have segment-level (strong) annotations and others have video-level (weak) annotations.}
    \label{tab: thumos_mix}
\end{table*}
\begin{table*}[ht]
    \centering
    \begin{tabular}{l|c|c|c|c|c|c|}
    Methods $\rightarrow$ & TRN~\cite{xu2019trn}& \multicolumn{5}{c|}{\textbf{WOAD}}\\
    Supervision $\rightarrow$ &S & V&V+$30\%$S&V+$50\%$S& V+$70\%$S& S \\
    \midrule
    mean F-AP $\rightarrow$ &69.1&66.7&66.9& 68.5& 69.3&\textbf{70.7} \\

    \end{tabular}
    \caption{Comparison with strongly-supervised SOTA method on ActivityNet1.2. V+$30\%$S means that $30\%$ of videos have segment-level (strong) annotations and others have video-level (weak) annotations.}
    \label{tab: anet_frame}
\end{table*}
\begin{table}[htbp]
    \centering
    \begin{tabular}{l|c||c|c}
        Methods & Sup. & mean F-AP & mean P-AP@ 1\\
        \midrule
        W/O RNN& \multirow{5}{*}{V}& 49.0& 19.7 \\
        W/O $\textbf{st}_i^t$ (Infer.)&  & 54.4&20.2 \\
        W/O temp. pool & &54.3 &21.6 \\
        \textbf{WOAD}& & \textbf{54.4}& \textbf{21.9} \\
        \midrule
        \midrule
        W/O $L_{TPG}$ & \multirow{5}{*}{S} & 61.2&24.5 \\
        W/O RNN & & 57.5& 24.6 \\
        W/O weak sup. &  &63.9 &25.4 \\
        W/O temp. pool & & 65.6&26.3 \\
        \textbf{WOAD} & & \textbf{67.1}& \textbf{28.0} \\
    \end{tabular}
    \caption{Ablation study of our proposed WOAD on THUMOS'14. V and S indicate video-level (weak) and segment-level (strong) supervision, respectively.}
    \label{tab: thumos_ablation}
\end{table}
\begin{table}[htbp]
    \centering
    \begin{tabular}{l|c||c|c}
        Methods & Sup. &mean F-AP & mean P-AP@ 1\\
        \midrule
         W-TALC~\cite{paul2018w} &\multirow{2}{*}{V} &48.0& 16.2 \\
        \textbf{WOAD} & &\textbf{54.4}& \textbf{21.9} \\
        \midrule
        \midrule
        BaS-Net~\cite{lee2020background} & \multirow{2}{*}{V}&51.0& 19.9 \\
        \textbf{WOAD}$+$ & &\textbf{57.8}& \textbf{22.7} \\
    \end{tabular}
    \caption{Comparison of our method implemented with different modules as TPG on THUMOS'14 dataset. \emph{WOAD}$+$ indicates our framework using BaS-Net as TPG.}
    \label{tab: offline_weak_detection}
\end{table}

\section{Experiments}
\textbf{Datasets}. We conduct experiments on THUMOS'14~\cite{THUMOS14}, ActivityNet1.2 and ActivityNet1.3~\cite{caba2015activitynet}. THUMOS'14 contains 20 sport-related action classes. Following prior works, we use the validation set (200 videos) for training and evaluate on the test set (212 videos). Each video contains 15 action instances on average. ActivityNet1.2 contains 100 action classes with an average of 1.5 action instances per video. We train on the training set (4819 videos) and evaluate on validation set (2383 videos). With 200 action classes, ActivityNet1.3 is an enlarged version of ActivityNet1.2. Our model is trained with the 10k training videos and validated using its 5k validation videos. Although ActivityNet datasets are much larger, THUMOS'14 has varying video lengths and much denser temporally annotated actions which make it more challenging.

\textbf{Evaluation metrics}. Following previous works~\cite{de2016online,gao2017red,xu2019trn,shou2018online,gao2019startnet}, frame-based average precision (F-AP) and point-based average precision (P-AP) are used as our evaluation metrics. F-AP focuses on evaluating model performance based on per-frame predictions. P-AP evaluates performance of action starts. P-AP works similarly as the bounding box based AP in the object detection task, except that P-AP uses time difference to determine whether an action start prediction is correct, while the later one uses Intersection over Union between the predicted box and the ground truth. The mean F-AP and mean P-AP are calculated by averaging F-APs and P-APs over classes, respectively.

\textbf{Baselines}. Since our TPG is implemented based on W-TALC~\cite{paul2018w}, we compare with this baseline to show the advantage of our framework under the weakly supervised setting.~\footnote{To perform fair comparison, we evaluate on its frame-level predictions ($\textbf{S}_i$ in Fig.~\ref{fig: pipeline}) during inference under the online constraint.} Then, we compare against two strongly-supervised methods, TRN~\cite{xu2019trn} and StartNet~\cite{gao2019startnet}. TRN is a state-of-the-art (SOTA) method for the online per-frame action recognition task and StartNet is the SOTA method for the task of online detection of action starts.

\textbf{Feature description}. On THUMOS'14 and ActivityNet1.2, two-stream (optical flow stream and RGB stream) I3D network~\cite{carreira2017quo} pre-trained on Kinetics is used as the feature extractor. Features are extracted at the chunk level. Video frames are extracted at 25 FPS and the chunk size is 16. The final features are the concatenation of the outputs of the two streams, resulting in a dimension of 2048. To perform fair comparison, our method and the baselines use the pre-extracted features provided by the authors of~\cite{paul2018w}. To avoid heavy feature extraction, we adopt the C3D features~\footnote{http://activity-net.org/challenges/2016/download.html} of ActivityNet1.3 officially released by the ActivityNet Challenge.

\textbf{Implementation details}. Our method is implemented using Pytorch. The update interval of temporal proposals is set to be $N=100$ for THUMOS'14, $N=500$ for ActivityNet1.2 and $1000$ for ActivityNet1.3. For OAR, the dimension of $\textbf{h}_i^t$ is set to be 4096 and the length of training sequence for LSTM is 64. $M$ in temporal pooling is fixed to be 3. $\gamma$ in Eq.~\ref{eq: oar_loss} is set to be 2. Since starts are sparsely located in each video, we use all positive frames and randomly sample 3 times negative ones in each training batch to compute start loss. $\lambda$ is fixed to be 0.5. Batch size of training videos is set to be 10. We use Adam~\cite{kingma2014adam} with weight decay $5\times10^{-4}$ and set learning rate to be $1\times10^{-4}$. 

\textbf{Supervision combination strategy}. When segment-level (strong) annotations exist, frame and start losses are computed using a combination of ground-truth and pseudo labels. The intuition is that the boundary annotations usually involve ambiguous decisions, so the noisy labels may serve as a type of regularization by making the label set reasonably diverse. We conduct the combination by randomly selecting $90\%$ videos using ground-truth supervision and other videos use the noisy proposal supervision. The proposals and the combination set are updated during training.

\begin{figure*}[t]
    \centering
    \includegraphics[width=0.84\linewidth]{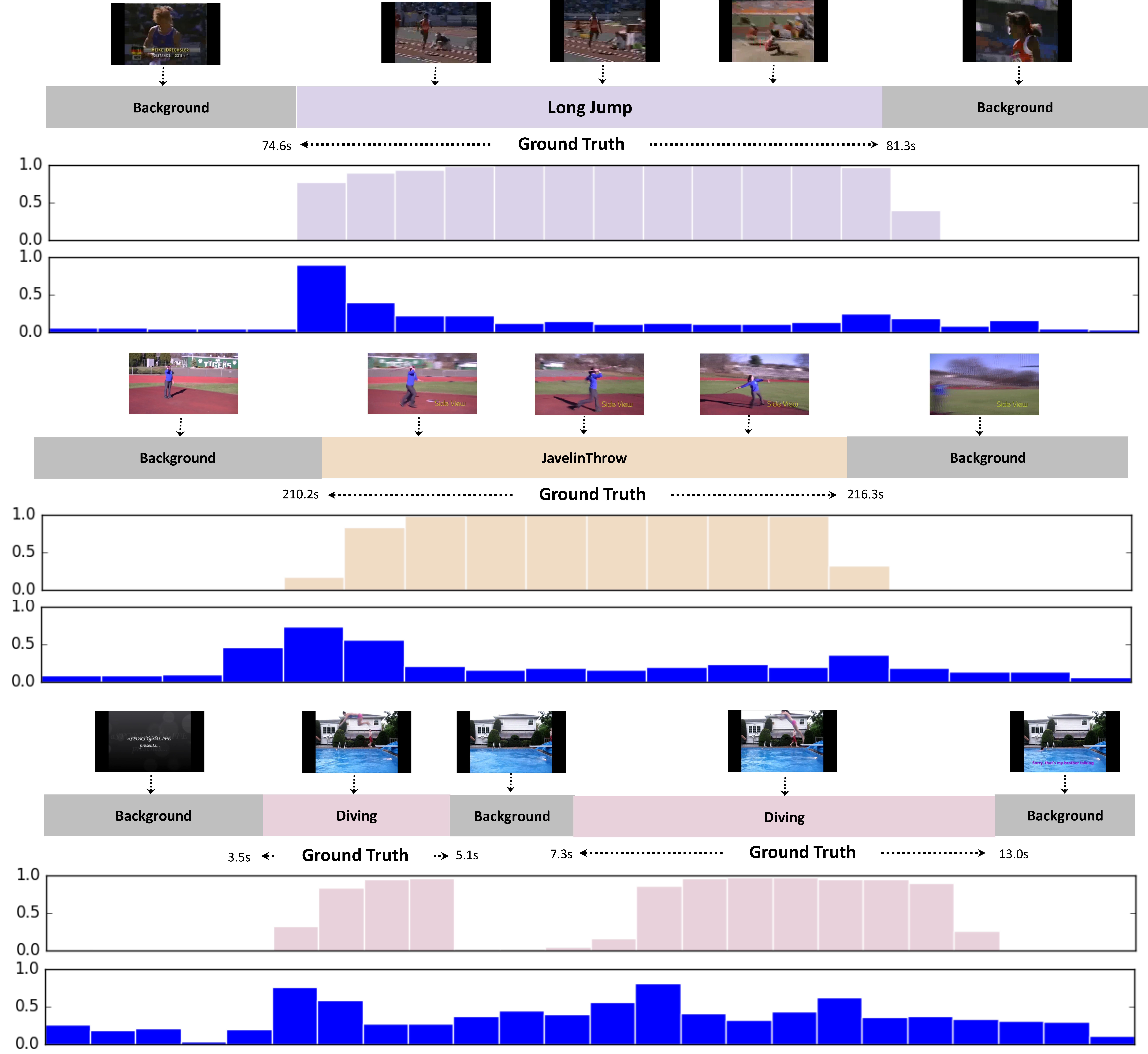}
    \caption{Qualitative results of our weakly-supervised method. The last row of each group indicates the predicted start scores (blue bars) and the second to the last row (colored bars) indicates the predicted action scores of the ground-truth class.
    }
    \label{fig: qualitative}
\end{figure*}

\subsection{Experimental Results}
\subsubsection{WOAD with Weak Supervision}
Our main focus is weakly supervised online action detection. So, we first conduct experiments under this setting.

\textbf{Online detection of action start}. Comparisons in terms of P-AP between our approach and our baseline, W-TALC, are shown in Table~\ref{tab: thumos_weak_compare} and~\ref{tab: activity1.2_weak_compare}. Our method outperforms W-TALC over all the time thresholds. Specifically, we obtain 5.7\% and 2.7\% higher mean P-AP when time threshold is 1 second on THUMOS'14 and ActivityNet1.2, respectively. 

When only using video-level annotations, our method obtains better performance than strongly-supervised StartNet in general on THUMOS'14, ActivityNet1.2 and ActivityNet1.3 as shown in Table~\ref{tab: thumos_start}, \ref{tab: anet_start} and~\ref{tab: anet_1.3}.

\textbf{Online per-frame action recognition}. Comparisons between our method and W-TALC in terms of F-AP are shown in Table~\ref{tab: thumos_weak_compare} and~\ref{tab: activity1.2_weak_compare} (last column). Our method improves the baseline largely by 6.4\% and 12.9\% mean F-AP on THUMOS'14 and ActivityNet1.2 which demonstrate the effectiveness of our framework.

We also compare our weakly-supervised method with strongly-supervised baselines in Table~\ref{tab: anet_1.3}, ~\ref{tab: thumos_frame} and~\ref{tab: anet_frame}. Our weakly-supervised method achieves $54.4\%$ mean F-AP improving the strongly-supervised TRN by $3.4\%$ on THUMOS'14, and obtains $66.7\%$ mean F-AP which is only $2.4\%$ lower than TRN on ActivityNet1.2. Although using video-level labels, our method achieves comparable results to strongly-supervised TRN. THUMOS'14 is a more challenging dataset, containing $10\times$ action instances per video as compared to ActivityNet, so it leaves more room for our model to improve the performance. This is why our method gains much better results on THUMOS'14. For future reference, our method obtains 67.9\% using weighted maIA, a newly proposed metric in~\cite{baptista2019rethinking}.

\subsubsection{WOAD with Strong Supervision}
\textbf{Full strong supervision}. When using all segment-level annotations, our method largely outperforms TRN on THUMOS'14 ( by $16.1\%$ mean F-AP, see Table~\ref{tab: thumos_mix}). On ActivityNet1.2 and ActivityNet1.3, our method achieves new state-of-the-art performance of $70.7\%$ mean F-AP (Table~\ref{tab: anet_frame}) and $46.8\%$ (Table~\ref{tab: anet_1.3}), respectively. For online detection of action start, our method outperforms StarNet consistently for all time thresholds on both THUMOS'14 (Table~\ref{tab: thumos_start}) and ActivityNet1.2 (Table~\ref{tab: anet_start}). Interestingly, we observe that the gap of our performance between strong- and weak-supervised settings is only 4\% and 2.8\% on ActivityNet1.2 and 1.3, whereas the gap is $12.7\%$ on THUMOS'14. This may be because that the average ratio of action length over video length in ActivityNet is $\sim40\%$, while the ratio is only $2\%$ in THUMOS'14. So, our method is not as sensitive to the boundary shift of a noisy proposal in ActivityNet as in THUMOS'14.

\textbf{WOAD with Mixed Supervision}. One advantage of our method is the flexibility of taking different forms of supervision for different videos. We evaluate our model when only a portion of randomly selected videos have segment-level annotations. As shown in Table~\ref{tab: thumos_mix} and~\ref{tab: anet_frame}, the performance of our model improves when more segment-level labels are available. On ActivityNet1.2, our method achieves comparable performance to previous SOTA method when only $70\%$ of data contains segment-level annotations.

\subsubsection{Model Ablation and Analysis}
\label{sec: ablation}
Our superior performance may attribute to (1) the improvements by jointly training TPG and OAR; (2) the effect of the supervision combination strategy and (3) our desirable structure. Ablation studies are conducted to analyze the effect of each component of WOAD. 

\textbf{TPG and OAR joint training}. Training two modules together has following benefits: (1) the diverse pseudo labels generated in different iterations could serve as regularization during training and (2) the shared features can be potentially improved by multi-task learning. We validate its effect by removing $L_{TPG}$ when strong labels are available. As shown in Table~\ref{tab: thumos_ablation}, disabling $L_{TPG}$ (\emph{W/O $L_{TPG}$}) results in $5.9\%$ and $3.5\%$ lower mean F-AP and mean P-AP@1. 

\textbf{Supervision combination}. Should we use the pseudo labels generated from TPG when strong labels are available? As shown in Table~\ref{tab: thumos_ablation}, using only segment-level supervision (\emph{W/O weak sup.}) results in degradation of mean F-AP by $3.2\%$ and mean P-AP@1 by $2.6\%$. We observe that small amount of pseudo labels could serve as a type of regularization thus relieve overfitting. However, adding too much noise would lead to performance degradation. For example, when we use pseudo labels for $90\%$ of videos, the mean F-AP is decreased to $58.0\%$.

\textbf{Effect of start point prediction}. As shown in Table~\ref{tab: thumos_ablation}, removing $st_i^t$ (\emph{W/O $\textbf{st}_i^t$ (Infer.)}) results in decreasing mean P-AP@1 by $1.7\%$. As expected, start point prediction improves the accuracy of action start generation by suppressing false positives at non-start frames. 

\textbf{Effect of temporal pooling}. Information of the current frame may not be the best indicator for start prediction, so we use temporal pooling to make our model more flexible to take temporal information. When it is removed, our model (\emph{W/O temp. pool} in Table~\ref{tab: thumos_ablation}) obtains worse performances. 

\textbf{Effect of RNN}. Our OAR utilizes LSTM to aggregate temporal information. To investigate the effect of the recurrent architecture, we replace the LSTM with two fully connected layers of size 4096. The performance (\emph{W/O RNN} in Table~\ref{tab: thumos_ablation}) under both weak and strong supervised settings are much worse than our method which demonstrate the usefulness of RNN in our architecture.

\textbf{Effect of $\lambda$}. The hyper parameter $\lambda$ in Eq.~\ref{eq: loss} controls the contribution of the losses from our TPG and OAR modules to the total loss. $\lambda$ is set to be 0.5 as default. Our method is relatively robust in this hyper-parameter choice. With video-level supervision, our method achieves 54.4\%, 55.0\% and 54.6\% mean F-AP when $\lambda$ equals 0.5, 1.0 and 2.0, respectively. When strongly supervised, our method obtains 67.1\%, 66.3\% and 66.6\% mean F-AP accordingly.

\textbf{WOAD with another TPG option}. Although, our TPG is based on~\cite{paul2018w}, other weakly supervised techniques may also serve as TPG. We experiment with another offline weakly supervised method, BaS-Net~\cite{lee2020background}, to generates action proposals. The results in Table~\ref{tab: offline_weak_detection} show that our method outperforms the baselines of \emph{W-TALC} and \emph{BaS-Net} largely by 6.4\% and 6.8\% mean F-AP, and by 5.7\% and 2.8\% mean P-AP@1, respectively. The clear gaps demonstrate the effectiveness of our design. Besides, WOAD$+$ performs much better than WOAD which suggests that our method achieves better results when using a more accurate TPG model. 

\textbf{Model efficiency analysis}.
Since our model and the baselines use the same features, we compare the inference times after feature extraction. We test all the models under the same environment with a single Tesla V100 GPU. The per-frame inference times of TRN, StartNet and our method averaging over the entire test set of THUMOS'14 are 2.60 ms, 0.56 ms and 0.40 ms respectively. The results suggests that our method is the fastest, around 6$\times$ faster than TRN. Model size is another key factor, especially for online tasks. Given similar model accuracy, smaller models are preferable, since they require less memory. Number of parameters of TRN, StartNet and our method (TPG+OAR) are 314M, 118M and 110M. Our method has the least number of parameters (3$\times$ smaller than TRN).

\textbf{Qualitative results}. In Fig.~\ref{fig: qualitative}, we visualize our predicted action and start scores of three representative cases. As it is shown, our method performs fairly good in the case of \emph{Long Jump}. In the second case (\emph{JavelinThrow}), our action scores are desirable. Since the visual appearance across frames near start points are very similar, the start scores are (although still reasonable) not as good as those in the first case. In the third case (\emph{Diving}), there are two actions occurred with a very short break in between. This makes it very hard to estimate starts based on the subtle visual and motion differences. Thus, the gap of start scores between start and non-start points are relatively small.

\textbf{Our weaknesses}. Our TPG is basically an offline weakly supervised method which performs poorly in long videos especially when there are very few training videos available. The low accuracy of the generated pseudo labels may result in an unsatisfactory performance of our method in the above scenarios. We conducted experiments on TVSeries~\cite{de2016online} dataset which contains only 20 long videos for training. With the two-stream features used in~\cite{gao2017red,xu2019trn}, our method achieves 59.1\% and 82.2\% mean cAP in weakly and strongly supervised settings. \cite{gao2017red} and~\cite{xu2019trn} reported 79.2\% and 83.7\%, respectively. As suggested by the results, our method is not as competitive as it is in other datasets.

\section{Conclusion}
We propose WOAD to address online action detection using weak supervision. Previous methods rely on segment-level annotations for training which leads to significant amount of human effort and hinders the model scalability. Our proposed WOAD can be trained using only video-level labels and is largely improved when strong labels are available. Experimental results demonstrate that our method with weak supervision obtains comparable performance to the existing strongly-supervised approaches on the online action detection tasks and achieves the state-of-the-art results when strongly supervised.

\noindent\textbf{Acknowledgement}.
Discussions with Peng Tang and Kathy Baxter are gratefully acknowledged. We thank Zuxuan Wu,
Zeyuan Chen and Salesforce researchers for the help of improving the writing.

{\small
\bibliographystyle{ieee_fullname}
\bibliography{egbib}
}

\end{document}